\pdfoutput=1
\PassOptionsToPackage{dvipsnames}{xcolor} 
\documentclass[11pt]{article}

\usepackage[]{acl}

\newcommand\blfootnote[1]{%
  \begingroup
  \renewcommand\thefootnote{}\footnote{#1}%
  \addtocounter{footnote}{-1}%
  \endgroup
}

\usepackage{times}
\usepackage{latexsym}
\usepackage[utf8]{inputenc}
\usepackage[T1]{fontenc}
\usepackage{CJKutf8}
\usepackage[english]{babel}
\usepackage{makecell}
\usepackage{graphicx}
\usepackage{hyperref}
\usepackage{multirow} 
\usepackage{adjustbox}
\usepackage{float} 
\usepackage{amsmath} 
\usepackage{adjustbox} 
\usepackage{verbatim} 
\usepackage{hwemoji}
\usepackage{pifont} 
\usepackage[normalem]{ulem}
\useunder{\uline}{\ul}{}
\usepackage[table,xcdraw]{}  
\usepackage{array}
\usepackage{algorithm}
\usepackage{algorithmicx}
\usepackage{algpseudocode}
\usepackage{dsfont}
\usepackage{tabularx}

\usepackage{listings}

\DeclareFontFamily{U}{mathx}{\hyphenchar\font45}
\DeclareFontShape{U}{mathx}{m}{n}{
      <5> <6> <7> <8> <9> <10>
      <10.95> <12> <14.4> <17.28> <20.74> <24.88>
      mathx10
      }{}
\DeclareSymbolFont{mathx}{U}{mathx}{m}{n}
\DeclareFontSubstitution{U}{mathx}{m}{n}
\DeclareMathAccent{\widebar}{0}{mathx}{"73}

\usepackage[T1]{fontenc}

\usepackage[utf8]{inputenc}

\usepackage{microtype}

\title{Beyond Single-User Dialogue: Assessing Multi-User Dialogue State Tracking Capabilities of Large Language Models}

\author{Sangmin Song, Juhwan Choi, JungMin Yun, \and YoungBin Kim  \\
  Chung-Ang University, Republic of Korea, Seoul \\
  \texttt{,@.com ,@.com \{cocoro357, ybkim85\}@cau.ac.kr} \\
}

\author{Sangmin Song\textsuperscript{1*}, Juhwan Choi\textsuperscript{2*}, JungMin Yun\textsuperscript{3}, \and YoungBin Kim\textsuperscript{3} \\
  \textsuperscript{1}GS Neotek \quad
  \textsuperscript{2}AITRICS \quad
  \textsuperscript{3}Chung-Ang University \quad \\
  \textsuperscript{1}\texttt{song0313@gsneotek.com} 
  \quad \textsuperscript{2}\texttt{jhchoi@aitrics.com} \\ \textsuperscript{3}\texttt{\{cocoro357, ybkim85\}@cau.ac.kr}
}

\begin{document}
\maketitle

\blfootnote{\textsuperscript{*}Equal contribution as co-first author.}

\begin{abstract}
Large language models (LLMs) have demonstrated remarkable performance in zero-shot dialogue state tracking (DST), reducing the need for task-specific training. However, conventional DST benchmarks primarily focus on structured user-agent conversations, failing to capture the complexities of real-world multi-user interactions. In this study, we assess the robustness of LLMs in multi-user DST while minimizing dataset construction costs. Inspired by recent advances in LLM-based data annotation, we extend an existing DST dataset by generating utterances of a second user based on speech act theory. Our methodology systematically incorporates a second user’s utterances into conversations, enabling a controlled evaluation of LLMs in multi-user settings. Experimental results reveal a significant performance drop compared to single-user DST, highlighting the limitations of current LLMs in extracting and tracking dialogue states amidst multiple speakers. Our findings emphasize the need for future research to enhance LLMs for multi-user DST scenarios, paving the way for more realistic and robust DST models.
\end{abstract}

\section{Introduction}
\label{sec:intro}

Large language models (LLMs) have achieved remarkable success across various natural language processing tasks, ranging from translation to mathematical problem-solving \cite{brown2020language, achiam2023gpt, dubey2024llama, team2024gemma, hurst2024gpt}. Their effectiveness has also been demonstrated in dialogue state tracking (DST), a task that predicts a user’s goal based on conversations between the user and an agent \cite{jacqmin2022you}. Traditional approaches to DST typically rely on training sequence-to-sequence models \cite{heck2020trippy, campagna2020zero, li2021zero}. However, recent studies suggest that LLMs can achieve competitive performance in DST through zero-shot inference, eliminating the need for task-specific training \cite{heck2023chatgpt}.

\begin{figure}[t]
    \centering
    \includegraphics[width=\columnwidth]{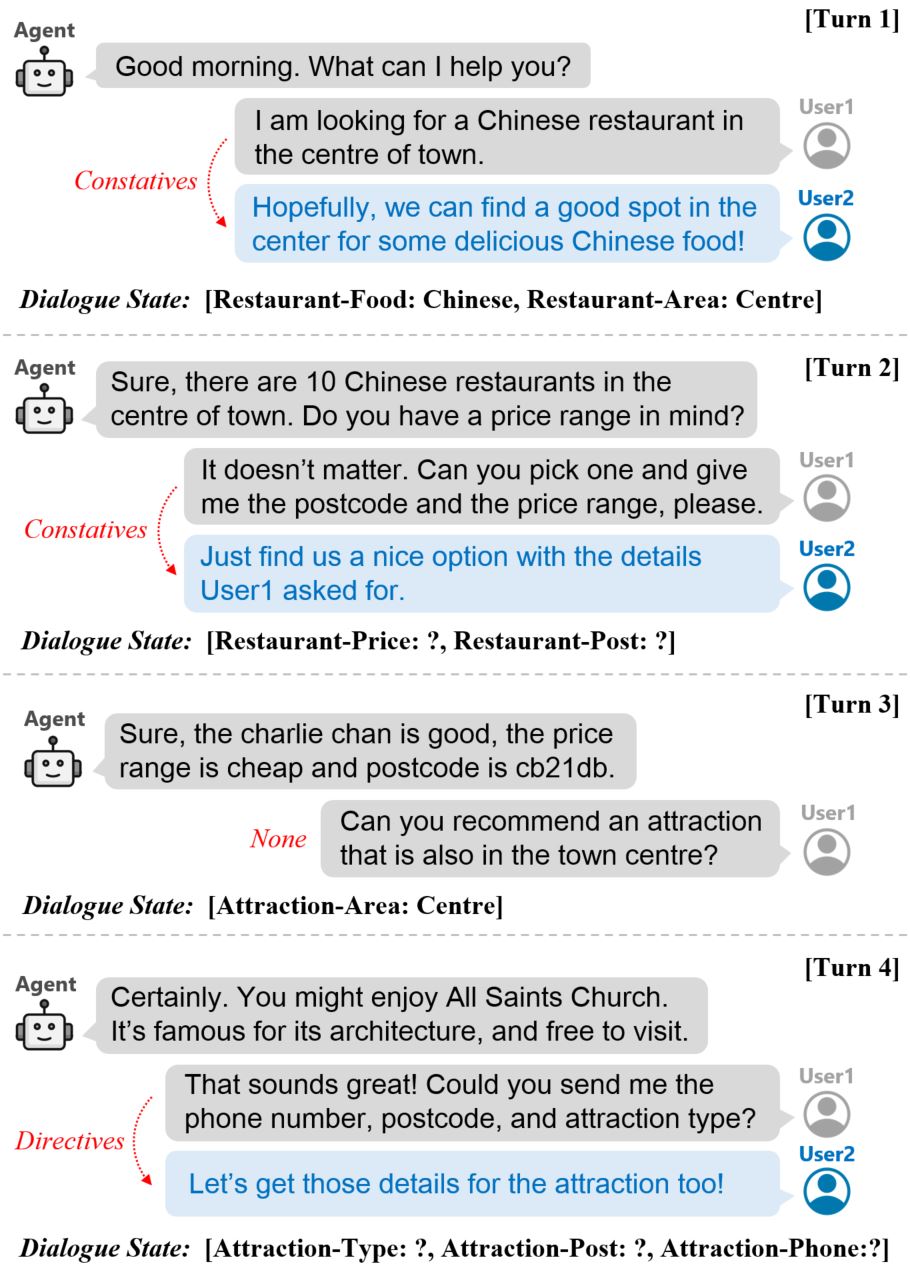}
    \caption{Example of a dialogue and slot values where utterance of user$_2$ is introduced through our proposed method. We use this multi-user dialogue to evaluate LLM-based zero-shot DST approaches and compare its performance with existing single-user DST.}
\label{fig:example}
\end{figure}

\begin{figure*}[t]
    \centering
    \includegraphics[width=\textwidth]{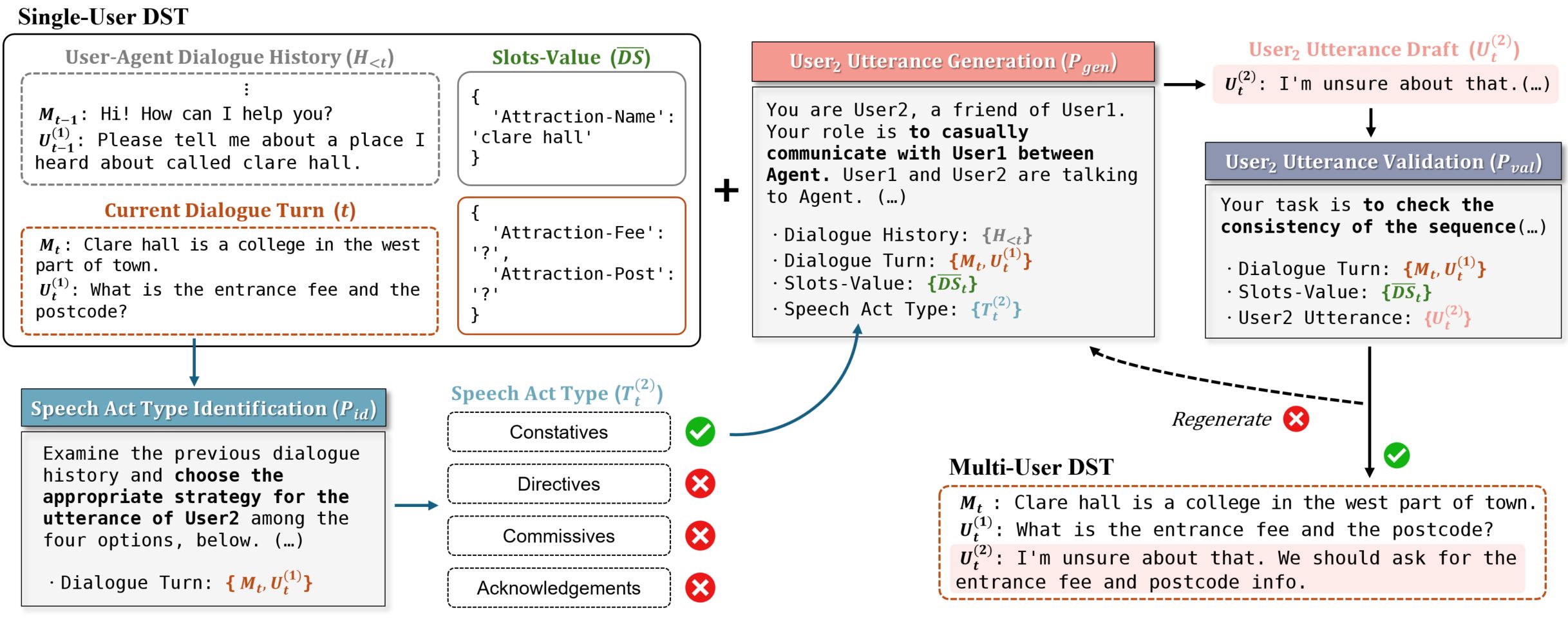}
    \caption{The illustration of our method stated in Section~\ref{sec:method-generation}.}
\label{fig:framework}
\end{figure*}

Despite these advances, conventional DST benchmarks primarily feature structured user-agent dialogues, limiting their applicability to more complex real-world scenarios. For instance, MultiWOZ \cite{budzianowski2018multiwoz}, one of the most widely used DST datasets, contains only single-user conversations. In contrast, real-world interactions often involve multiple users collaboratively engaging in decision-making before interacting with an agent. For example, a group of friends discussing restaurant options might first debate their cuisine preferences and locations before making a request to a booking system (e.g., "I prefer Korean food." → "Is there a good Korean restaurant in Karlsplatz, Vienna?"). Assessing DST models in multi-user scenarios is therefore essential for evaluating their robustness and generalizability.

In this work, we conduct a case study to assess the performance of LLMs in multi-user DST, with minimal dataset construction costs. Inspired by recent efforts to employ LLMs for data annotation, we extend existing DST datasets using state-of-the-art LLMs \cite{tan2024large, choi2024multi}. Specifically, we generate additional user utterances that interleave between the original user-agent exchanges. Our approach comprises two main steps: (1) identifying the appropriate type of interjections based on speech act theory \cite{austin1975things, bach1979linguistic}, and (2) prompting an LLM to generate and validate the second user’s utterances accordingly. Figure~\ref{fig:example} presents an example of a dialogue with utterances of the second user, based on our suggested framework. We compare the performance of LLMs in cases where the second user utterance is present or absent throughout the experiment.

Experimental results on our extended dataset show a substantial performance drop compared to single-user DST. This indicates that current LLMs struggle to correctly extract and track dialogue states in conversations with multiple users, underscoring limitations in their generalizability. Our findings highlight the need for future research to develop DST methods and prompt-engineering strategies optimized for multi-user settings.

Our primary contributions are as follows:
\begin{itemize}
    \item We propose the first systematic evaluation of LLMs in multi-user DST, highlighting the unique challenges posed by multi-user interactions.
    \item We introduce a cost-efficient method to extend existing DST datasets by generating additional user utterances based on speech acts using state-of-the-art LLMs.
    \item We conduct comprehensive experiments demonstrating a notable performance drop in multi-user DST, underscoring the limitations of current LLMs when handling complex dialogue dynamics.
\end{itemize}

\section{Methodology}
\label{sec:method}
\subsection{Preliminaries}
\label{sec:method-preliminary}

In this subsection, we briefly explain preliminaries regarding DST, especially focusing on zero-shot DST using LLMs. We follow the definitions and notations of a previous study \cite{heck2023chatgpt}.

DST is a core component of task-oriented dialogue systems, responsible for tracking the user’s evolving intent throughout a conversation. At each dialogue turn $t$, given a user utterance $U_t$ and a preceding agent utterance $M_t$, DST predicts a structured set of slot-value pairs representing the user’s goal. We call this structured set of slot-value pairs the dialogue state $\textit{DS}_t$, which encapsulates the user’s intent and relevant context at turn $t$. $\textit{DS}_t$ is updated as follows:

\begin{gather*}
\textit{DS}_t = \textit{DS}_{t-1} \cup \widehat{\textit{DS}}_t
\end{gather*}
where $\widehat{\textit{DS}}_t$ is the newly inferred slot-value set at turn $t$.

Traditional DST models require extensive supervised training on labeled datasets \cite{heck2020trippy, campagna2020zero, li2021zero}. However, recent advances in LLMs have enabled zero-shot DST, where a general-purpose model extracts slot values without explicit fine-tuning \cite{feng2023towards, heck2023chatgpt}. In zero-shot DST, the inference-time slot set \( S = \{(s_1, v_1), (s_2, v_2), ..., (s_n, v_n)\} \), where \( s_i \) denotes a slot and \( v_i \) its corresponding value, is disjoint from the training-time slot set \( S' \), meaning \( S \cap S' = \emptyset \). This requires the model to generalize without prior domain-specific training.

Zero-shot DST using LLMs typically follows a prompting-based approach. $P$, an instruction including task description, is provided in natural language, guiding the model to extract relevant slot-value pairs from a conversation. At each turn, the prompt $A_t$ includes both agent and user utterances:

\begin{gather*}
A_1 = P \oplus \text{"agent": } M_1 \oplus \text{"user1": } U^{(1)}_1, \\
A_t = \text{"agent": } M_t \oplus \text{"user1": } U^{(1)}_t, \quad \forall t \in [2, T]
\end{gather*}

Unlike traditional approaches, LLMs perform DST by leveraging their conversational alignment and pre-trained world knowledge without modifying its parameters. In this study, we aim to validate the capabilities of LLMs in handling multi-user DST scenario.

\subsection{Construction of Multi-user DST Dataset}
\label{sec:method-generation}

This subsection outlines our method to extend the existing dataset to incorporate a second user, thereby establishing multi-user DST.  To accomplish this, we follow the concept of speech act theory \cite{austin1975things, bach1979linguistic}, which defines the types of human utterances.

We adopt four speech act types based on an established study \cite{bach1979linguistic}:

\begin{itemize}
    \item \textbf{Constatives}: Statements that express factual information or claims, including answering, confirming, denying, disagreeing, or stating.
    \item \textbf{Directives}: Attempts to influence the actions of the addressee, such as advising, asking, requesting, forbidding, inviting, or ordering.
    \item \textbf{Commissives}: Utterances that commit the speaker to a future action, including promising, planning, vowing, betting, or opposing.
    \item \textbf{Acknowledgments}: Expressions of the speaker’s attitude toward the hearer’s social actions, such as apologizing, greeting, thanking, or accepting acknowledgments.
\end{itemize}

Based on these speech act types, we introduce user$_2$ utterances ($U^{(2)}_{t}$) into existing user-agent dialogues. Specifically, we insert $U^{(2)}_{t}$ at each dialogue turn $t$, within the original user$1$ utterance ($U^{(1)}_{t}$) and the agent response ($M_{t}$). To achieve this, we first use an LLM to determine the appropriate speech act type for $U^{(2)}_{t}$. This classification is represented as $T^{(2)}_t=\mathcal{L}(\mathcal{P}_{\textit{id}}, U^{(1)}_{t}, M_{t})$, where $\mathcal{L}$ denotes the LLM and $\mathcal{P}_{\textit{id}}$ represents the corresponding prompt.

Once the speech act type is identified, we generate $U^{(2)}_{t}$ by prompting the LLM with relevant context, including $T^{(2)}_t$, $U^{(1)}_{t}$, $M_{t}$, dialogue history ($H_{<t}$), and the gold slot value ($\widebar{\textit{DS}}_t$). This process is formulated as $U^{(2)}_{t} = \mathcal{L}(\mathcal{P}_{\textit{gen}}(T^{(2)}_t), H_{<t}, U^{(1)}_{t}, M_{t}, \widebar{\textit{DS}}_t)$.

To ensure the generated $U^{(2)}_{t}$ maintains conversation coherence and does not introduce hallucinated slot values, we conduct a validation step. If $V_t = \mathcal{L}(\mathcal{P}_{\textit{val}}, U^{(1)}_{t}, U^{(2)}_{t}, M_{t}, \widebar{\textit{DS}}_t)$ returns True, we accept $U^{(2)}_{t}$ and insert it into the dialogue. Otherwise, we regenerate and revalidate $U^{(2)}_{t}$ up to three times. If validation fails in all attempts, we discard $U^{(2)}_{t}$ by setting it to $\emptyset$, ensuring that its inclusion does not compromise dialogue integrity.

This procedure is applied iteratively across all dialogue turns, extending the dataset to a multi-user DST setting. When performing DST on the extended dataset, we append user$_2$’s utterance to the prompt $A_t$ as follows:

\begin{gather*}
A_t' = A_t \oplus \text{"user2": } U_t^{(2)} \thinspace\thinspace \text{ if } U_t^{(2)} \thinspace \neq \emptyset,\thinspace\thinspace\thinspace \forall t \in [1, T]
\end{gather*}

\section{Experiment}
\label{sec:experiment}

\subsection{Experimental Design}
\label{sec:experiment-evaluation}

We introduce our experimental design to assess the performance of LLMs on multi-user DST using the extended dataset. 

\paragraph{Evaluation Models.} We adopt five different LLMs for experiments: GPT-4o \cite{hurst2024gpt}, Claude 3.5 Sonnet \cite{anthropic2024claude35sonnet}, Gemini-2.0-Flash \cite{pichai2024gemini}, LLaMA-3.1-8B \cite{dubey2024llama}, and Gemma-2-9B \cite{team2024gemma}. These models are selected to evaluate performance across different pretraining paradigms and parameter sizes. Further details can be found in Appendix~\ref{app:impl-details}.

\paragraph{Zero-Shot DST Setup.} We use a prompting-based approach following \cite{heck2023chatgpt}. The prompt provides a task description, slot definitions, and the current dialogue history. At each turn, models extract slot-value pairs based on their internal knowledge. We keep inference parameters (temperature, top-p) at default values to ensure consistency across all models. A detailed breakdown of the prompt format is provided in Appendix~\ref{app:prompt-experiment}.

\paragraph{Evaluation Metric.} To assess the performance of each model, we use Joint Goal Accuracy (JGA) \cite{henderson2014second}, which measures whether all slot values at each turn match the ground truth:
\begin{equation}
\textit{JGA} = \frac{1}{N} \sum^{N}_{i=1}\mathds{1}(\widehat{\textit{DS}}_t=\widebar{\textit{DS}}_t)
\end{equation}
where $N$ is the total number of turns, $\widehat{\textit{DS}}_t$ is the predicted dialogue state, and $\widebar{\textit{DS}}_t$ is the ground truth. We measure JGA of each model on the original MultiWOZ 2.1 \cite{eric2020multiwoz} and its extended version based on our methodology as stated in Section~\ref{sec:method-generation}, thereby comparing the performance of the LLM on single-user and multi-user scenario. Please refer to Appendix~\ref{app:dataset-stats} for the statistics of the extended dataset. By analyzing the performance gap between these settings, we quantify the impact of the existence of $U^{(2)}_{t}$ on DST performance. Specifically, we hypothesize that LLMs will show a performance drop in a multi-user DST, even if the topic, context, and slot value of the dialogue remain identical.

\subsection{Result}
\label{sec:experiment-result}

\begin{table}[t]
    \centering
    \resizebox{0.95\columnwidth}{!}{
    \begin{tabular}{l|ccccc|c}
        \Xhline{3\arrayrulewidth}
        Models & Attr. & Hotel & Rest. & Taxi & Train & Avg. \\  
        \hline\hline
        GPT-4o            & 56.8 & 46.0 & 55.1 & 69.3 & 61.9 & 57.82 \\
        w/ user$_2$ utterances     & -4.6 & -2.1 & -2.2 & -1.7 & -7.1 & -3.54 \\
        \hline
        Claude 3.5 Sonnet & 64.9 & 47.5 & 55.8 & 68.2 & 74.0 & 62.08 \\
        w/ user$_2$ utterances     & -3.8 & -1.9 & -2.6 & -1.3 & -6.7 & -3.26 \\
        \hline
        Gemini-2.0-Flash  & 36.7 & 25.3 & 24.4 & 61.9 & 28.0 & 35.26 \\
        w/ user$_2$ utterances     & -2.2 & -2.5 & -3.1 & -1.2 & -2.6 & -2.32 \\
        \hline
        LLaMA-3.1-8B      & 25.4 & 23.2 & 18.8 & 46.9 & 25.5 & 27.96 \\
        w/ user$_2$ utterances     & -1.6 & -0.3 & -1.2 & -0.4 & -1.8 & -1.06 \\
        \hline
        Gemma-2-9B        & 40.0 & 36.3 & 44.4 & 61.5 & 35.4 & 43.52 \\
        w/ user$_2$ utterances     & -1.1 & -0.2 & -2.0 & -1.1 & -2.4 & -1.36 \\
        \Xhline{3\arrayrulewidth}
    \end{tabular}
    }
    \caption{Performance comparison between LLM models for zero-shot DST in per-domain JGA. The first row of each model represents the performance of the model on the original dataset (i.e., no user$_2$ utterances), and the second row denotes the performance degradation of the model on the extended dataset with $U^{(2)}_{t}$ compared to the original dataset. Note that Attr. and Rest. denote attraction and restaurant, respectively. Please refer to Appendix~\ref{app:slot-eval} for slot-level evaluation.}
    \label{tab:performance}
\end{table}

Table~\ref{tab:performance} presents a detailed performance comparison of various LLMs under both single-user and extended multi-user DST settings. Across all evaluated models, we observe a consistent drop in JGA when additional user$_2$ utterances are present, thus confirming our hypothesis that the presence of multiple speakers introduces added complexity. In particular, models such as GPT-4o and Claude 3.5 Sonnet experience more severe degradations (e.g., up to -7.1\% in the train domain), indicating that even state-of-the-art LLMs struggle to disentangle the intertwined dialogue cues from multiple users.

Our qualitative analysis in Appendix~\ref{app:error-analysis} attributes these drops to two main error types: (1) mis-extraction of slot values when user$_2$ utterances distract the model and (2) incorrect inference of slot types due to unintended slot predictions. These findings emphasize the need for more robust, speaker-aware DST strategies. Future work should explore sophisticated prompt engineering or model architectures to distinguish multiple speakers.

\section{Conclusion}
\label{sec:conclusion}

We presented a systematic investigation of multi-user DST, a task that has been largely overlooked in traditional DST research. By extending an existing dataset through LLM-based generation of additional user$_2$ utterances, we revealed that contemporary LLMs face notable performance drops in multi-user DST scenarios, even when the underlying user request remains unchanged. Our findings underscore the importance of focusing on multi-user dialogue and highlight the challenges in maintaining accurate slot tracking when multiple speakers contribute to the conversation. 

Future work could explore more advanced prompt-engineering strategies or model designs that disentangle each speaker’s goal. In real-world applications, multiple people often share or refine a collective plan, such as making group travel arrangements or ordering meals. Improving LLM-based DST for these more realistic, collaborative settings will be crucial for developing robust, universally applicable dialogue systems.

\newpage
\section*{Limitations}

While our methodology offers a straightforward means to extend a single-user dataset to a multi-user dialogue scenario, it does not fully capture the dynamic and often collaborative nature of real-world multi-user interactions. Below, we elaborate on the key limitations:

\paragraph{Limited Conversation Realistic and Dynamics.} 
Our approach places additional speaker utterances into existing user-agent conversations without substantially altering the dialogue flow or the user’s underlying intent, to accurately evaluate the effect of the additional utterance. In realistic multi-user settings, however, users often negotiate, question, and refine each other’s ideas in ways that can change the system’s ultimate goals. For example, two users might disagree on a restaurant’s location or cuisine. Our current method does not model these forms of dynamic interaction, potentially less realistic and underrepresenting more complex use cases. Nonetheless, it is important to note that current relatively minor additions of multi-user interactions bring significant performance degradation, which suggests that more realistic interactions would likely pose even greater challenges, further highlighting the need for research in this field.

\paragraph{Quality of Generated Utterances.}
We rely on LLMs to generate and validate the additional user’s utterances. Although we employ validation steps to filter out incoherent or hallucinatory utterances, the generated data may still not fully reflect natural human conversational patterns. Furthermore, because these inserted utterances are not human-annotated, there is a risk that subtle contextual nuances are lost, and some inadvertently introduced biases or errors could remain unchecked. To precisely examine the quality of the generated utterances, we performed a manual inspection across three dimensions, which is presented in Appendix~\ref{app:verify-data-quality}.

\paragraph{Focus on Zero-shot DST.} 
We specifically target zero-shot DST based on LLMs to highlight the inherent challenge posed by multi-user DST. While this isolates the effect of additional user utterances, it also may not reflect performance under scenarios where LLMs are not used. However, from our initial attempt to conduct experiments using existing DST methods that do not use LLMs, we found that were not capable of handling this new multi-user interaction structure effectively. Accordingly, future research could investigate whether non-LLM or fully fine-tuned approaches better mitigate these multi-user complexities.

Overall, while our study demonstrates the feasibility of extending single-user dialogue datasets to multi-user scenarios at low cost, it serves primarily as a proof of concept to validate the limitation of current LLMs for multi-user DST, which is the key objective of this paper. Further methodological refinements in future studies, including human validation, richer conversation modeling, and domain diversification, will be essential to capture the true depth and breadth of multi-user task-oriented dialogues.



\bibliography{custom}
\bibliographystyle{acl_natbib}

\appendix

\section{Dataset Statistics}
\label{app:dataset-stats}

\begin{figure}[t]
    \centering
    \includegraphics[width=\columnwidth]{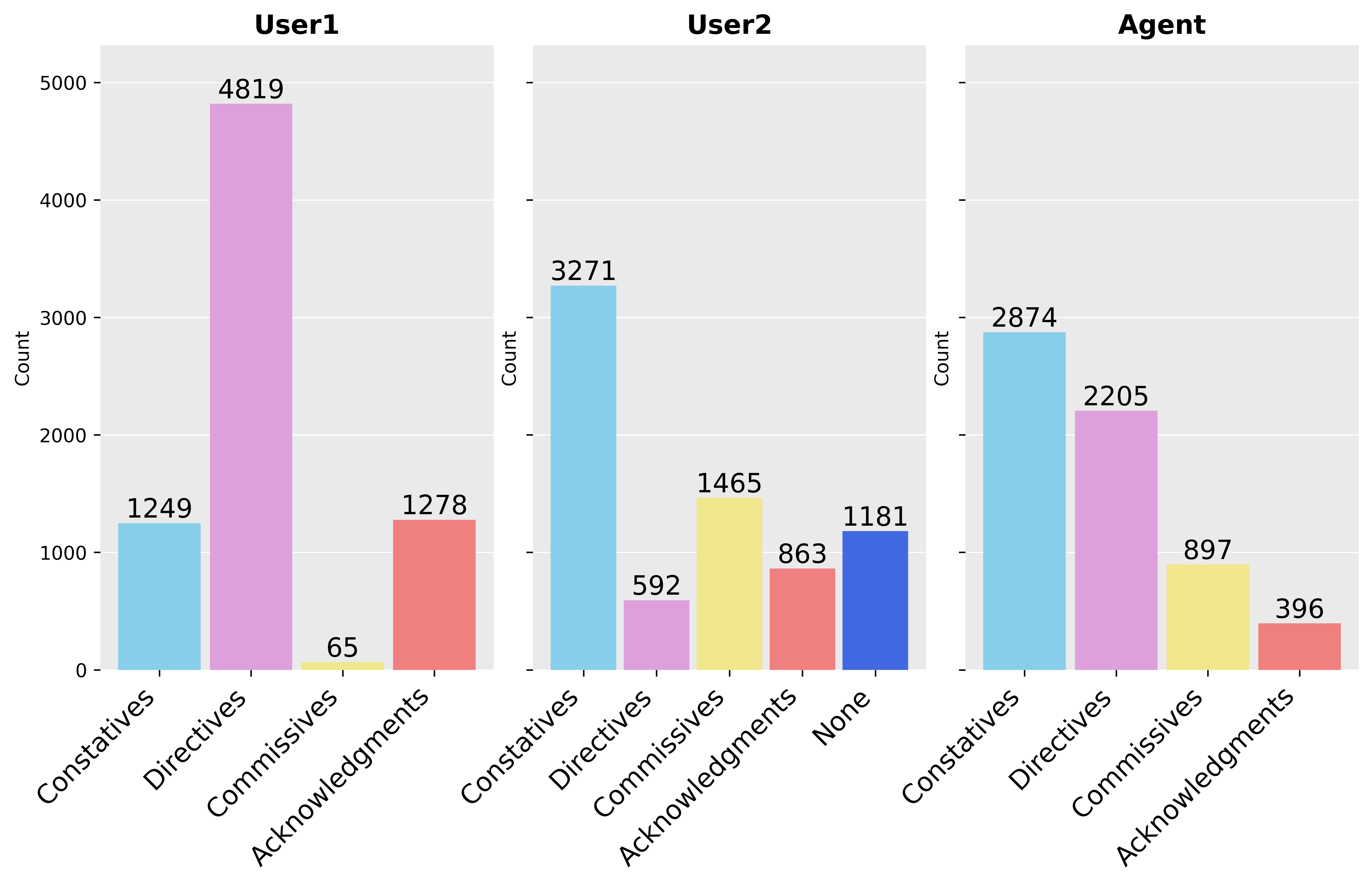}
    \caption{Distribution of speech acts of User$_1$, User$_2$, and Agent. We used our speech act type identification method stated in Section~\ref{sec:method-generation} to classify the speech act type of User$_1$ and Agent for comparison with User$_2$.}
\label{fig:type-distribution}
\end{figure}


In this section, we provide a detailed analysis of the dataset constructed for evaluating multi-user DST based on the aforementioned methodology. Our dataset is derived from the test set of MultiWOZ 2.1 \cite{eric2020multiwoz}, an existing DST benchmark. Accordingly, the dataset consists of 1,000 conversations with an average of 7.36 turns per dialogue, covering five different domains: attraction, hotel, restaurant, taxi, and train.

To examine the characteristics of multi-user interactions, we analyze the distribution of speech acts of $U^{(2)}_{t}$ within the dataset, which is depicted in Figure~\ref{fig:type-distribution}. Among all user$_2$ utterances, 44\% are constatives, 8\% are directives, 20\% are commissives, 12\% are acknowledgments, and 16\% are none (i.e., no $U^{(2)}_{t}$ at $t$).

Next, we compare the length of $U^{(2)}_{t}$ and $U^{(1)}_{t}$ in our dataset. On average, original utterances of $U^{(1)}_{t}$ contain 11.73 words, while the newly generated $U^{(2)}_{t}$ are slightly shorter, averaging 10.26 words. This difference arises because $U^{(2)}_{t}$ often consists of clarifications or brief confirmations rather than full-length requests or responses. $M_{t}$ tends to be longer, averaging 16.05 words, as they provide detailed responses or requests for clarification.

\section{Implementation Details}
\label{app:impl-details}

This section outlines implementation details to reproduce the experimental findings of our study. First, we used GPT-4o-2024-08-06 as $\mathcal{L}$ to extend the MultiWOZ 2.1 dataset, following the procedure stated in Section~\ref{sec:method-generation}. Second, for the experiment in Section~\ref{sec:experiment}, we used GPT-4o-2024-08-06, claude-3-5-sonnet-20241022, gemini-2.0-flash-001, Meta-Llama-3-8B-Instruct, and gemma-2-9b-it for each model. For hyperparameters such as temperature and top-p, we followed the default setup of each model. 

\section{Slot-Level Evaluation}
\label{app:slot-eval}

In this section, we present the experimental result measured by slot-level accuracy metrics, to supplement Section~\ref{sec:experiment-result}, which used JGA as evaluation metric \cite{dey2022towards, wang2022stop, yang2023multi, luo2024zero}. In this evaluation, we used following metrics:

\begin{itemize}
    \item \textbf{Accuracy — "None"}: Accuracy when the ground truth is "None" and the model correctly predicts "None".
    \item \textbf{Accuracy — "Dontcare"}: Accuracy when the user indicates no preference ("dontcare") and the model correctly captures it.
    \item \textbf{Accuracy — "Copy\_value"}: Accuracy when the model correctly copies a value from the user’s utterance.
    \item \textbf{Accuracy — "True"}: Accuracy when the ground truth is "true" and the model predicts "true".
    \item \textbf{Accuracy — "False"}: Accuracy when the ground truth is "false" and the model predicts "false".
    \item \textbf{Accuracy — "Refer"}: Accuracy in correctly identifying references to previously mentioned slot values.
    \item \textbf{Accuracy — "Inform"}: Accuracy in correctly predicting values explicitly provided (informed) by the user.
    \item \textbf{Slot Accuracy (SA)}: Slot Accuracy evaluates the model’s ability to correctly predict each individual (domain, slot, value) triplet at every dialogue turn. Unlike JGA, which requires the entire set of predicted belief states to exactly match the ground truth, SA compares each slot prediction independently. A prediction is counted as correct only if the domain, slot name, and slot value all match the ground truth \cite{wu2019transferable}.
\end{itemize}

The results of this analysis are presented in Table~\ref{tab:performance-slot}. They demonstrate that our findings in JGA are consistent with the slot-level metrics, which provide a more fine-grained evaluation compared to JGA. These metrics also reveal the same trend discussed in Section~\ref{sec:experiment-result}: models such as GPT-4o and Claude 3.5 Sonnet experience more pronounced performance degradation, further supporting our observations.

\begin{table*}[]
\centering
    \resizebox{\textwidth}{!}{
\begin{tabular}{l|ccccccc|cc}
\Xhline{3\arrayrulewidth}
Models                 & "None" & "Dontcare" & "Copy\_value" & "True" & "False" & "Refer" & "Inform" & SA  & JGA  \\ \hline\hline
GPT-4o                 & 94.8   & 74.6       & 90.6          & 96.6   & 89.7    & 7.8     & 64.0     & 94.4 & 57.82 \\
w/ user$_2$ utterances & -3.4   & -29.3      & -40.6         & -27.8  & -22.8   & -0.1    & -25.8    & -5.2 & -3.54 \\ \hline
Claude 3.5 Sonnet      & 96.2   & 75.9       & 93.6          & 66.8   & 87.6    & 9.7     & 63.3     & 95.7 & 62.08 \\
w/ user$_2$ utterances & -0.2   & -2.2       & -0.3          & -0.7   & -10.7   & -1.3    & -0.8     & -3.3 & -3.26 \\ \hline
Gemini-2.0-Flash       & 88.2   & 4.7        & 20.7          & 26.7   & 23.4    & 0.8     & 19.2     & 84.8 & 35.26 \\
w/ user$_2$ utterances & -0.2   & -0.2       & -0.3          & -2.7   & -0.6    & -0.3    & -0.3     & -0.2 & -2.32 \\ \hline
LLaMA-3.1-8B           & 86.0   & 27.6       & 29.1          & 39.4   & 29.0    & 5.9     & 23.4     & 83.1 & 27.96 \\
w/ user$_2$ utterances & -0.8   & -17.3      & -8.1          & -12.2  & -6.9    & -0.1    & -7.0     & -1.1 & -1.06 \\ \hline
Gemma-2-9B             & 91.3   & 34.5       & 74.8          & 60.8   & 57.2    & 15.4    & 52.8     & 90.2 & 43.52 \\
w/ user$_2$ utterances & -0.7   & -1.7       & -2.4          & -0.6   & -2.8    & -2.5    & -2.1     & -0.8 & -1.36 \\ \Xhline{3\arrayrulewidth}
\end{tabular}}
\caption{The performance of each LLM models for zero-shot DST in slot-level metrics.}
\label{tab:performance-slot}
\end{table*}

\section{Verification of Data Quality}
\label{app:verify-data-quality}
\subsection{Human Evaluation}

In this paper, we introduced a novel method for incorporating a second user’s utterance to simulate a multi-user dialogue environment. This section outlines the manual evaluation conducted to assess the quality of the generated utterances and the dataset constructed using our approach.

To perform this evaluation, we manually reviewed 100 dialogues—representing 10\% of the total dataset—based on specific criteria. Three evaluators, all volunteers from the authors’ research group, were tasked with assessing the selected dialogues according to the following dimensions:

\begin{itemize}
\item \textbf{Absence of Bias}: Identification of biased or inappropriate content in user$_2$ utterances.
\item \textbf{Utterance Quality}: Assessment of fluency and coherence within the dialogue context.
\item \textbf{Slot Consistency}: Verification that introducing user$_2$ does not inadvertently alter slot values, which would introduce noise.
\end{itemize}

The results of this evaluation are summarized in Table~\ref{tab:human-eval}. First, no instances of biased or inappropriate content were found in the reviewed dialogues, confirming that user$_2$ utterances consistently maintained a neutral and respectful tone. We attribute this to the controlled generation process, where user$_2$ responses are tightly constrained by the context provided by user$_1$.

Second, 92.68\% of user$_2$ utterances were judged as natural and contextually appropriate, indicating that our generation method effectively preserves dialogue coherence. The remaining cases exhibited minor awkward phrasing or slight tonal mismatches, but these did not significantly hinder comprehension.

Third, slot value consistency was maintained in 94.51\% of the dialogues, suggesting that the inclusion of user$_2$ rarely introduced unintended semantic shifts. In the few inconsistent cases, the discrepancies were due to subtle rephrasings that altered the slot interpretation slightly, rather than clear-cut errors. Notably, this consistency rate reflects a deliberately strict evaluation policy: evaluators were instructed to mark a slot as inconsistent whenever ambiguity was present, regardless of dataset constraints.

For example, consider the following exchange:

\begin{itemize}
\item \textbf{User1}: "I need to find a hotel that has free parking."
\item \textbf{User2}: "Make sure the hotel offers free parking for both of us."
\item \textbf{Slot}: hotel-parking = Yes
\end{itemize}

In this case, the slot was marked as inconsistent, based on the interpretation that user$_2$ implicitly requests an additional parking space. However, in MultiWOZ 2.1, the \texttt{hotel-parking} slot is binary (e.g., \texttt{yes}/\texttt{no}), and the model is not expected to distinguish between single or multiple parking requests. Thus, while human annotators may perceive this as inconsistent, such differences are typically irrelevant to the DST model. In this sense, the reported 5.49\% inconsistency rate should be viewed as an upper bound, reflecting a conservative human-centric judgment rather than an estimate of likely performance degradation (e.g., in JGA).

\begin{table}[]
\centering
    \resizebox{\columnwidth}{!}{
\begin{tabular}{l|ccc}
\Xhline{3\arrayrulewidth}
           & Absence of Bias & Utterance Quality & Slot Consistency \\ \hline
Ratio (\%) & 100.00          & 92.68             & 94.51           \\ \Xhline{3\arrayrulewidth}
\end{tabular}
}
\caption{Results of the human evaluation. A higher ratio indicates better performance.}
\label{tab:human-eval}
\end{table}

\subsection{Ablation Study on Clean Dialogues}

In addition to the human evaluation, we conducted an ablation study using 100 dialogues that had been flagged by evaluators as containing no slot inconsistencies. This experiment was carried out using GPT-4o, and the results are presented in Table~\ref{tab:performance-clean}. The findings reveal a similar pattern of performance degradation, even in these clean dialogues. This suggests that the minor inconsistencies identified during manual inspection do not significantly affect the overall conclusions of the study.

\begin{table}[]
\centering
    \resizebox{\columnwidth}{!}{
\begin{tabular}{l|ccccc|c}
\Xhline{3\arrayrulewidth}
                       & Attr. & Hotel & Rest. & Taxi & Train & Avg. \\ \hline \hline
GPT-4o                 & 39.4  & 43.4  & 60.9  & 62.8 & 62.6 & 53.82 \\
w/ user$_2$ utterances & -7.4  & -2.6  & -2.2  & -0.5 & -5.3 & -3.60 \\ \Xhline{3\arrayrulewidth}
\end{tabular}
}
\caption{Performance of GPT-4o on clean dialogues without inconsistent slots.}
\label{tab:performance-clean}
\end{table}

\section{Error Analysis}
\label{app:error-analysis}

This section provides an analysis of the errors induced by the introduction of user$_2$ utterances in multi-user DST. We manually investigated cases where the LLM correctly inferred slot values in the original dataset but misclassified them when the dataset was extended with additional user utterances. Errors primarily fall into two categories: (1) errors in extracting the correct slot value and (2) errors in identifying the correct slot type. Below, we provide examples and discussions for each type.

Table~\ref{tab:error-analysis-1} shows a conversation where the user asks for information about "Warkworth House." In the single-user scenario, the LLM correctly identifies "Warkworth House" as the value of the \textit{Hotel-Name} slot. However, once an additional user$_2$ utterance is introduced (expressing curiosity about the history of Warkworth House), the model fails to track the intended hotel name. This suggests that the extra mention of historical information deflects the model’s attention, causing it to overlook the slot value relevant to user$_1$’s primary request. Importantly, the problem seems to stem from the model’s difficulty in separating the primary user request (i.e., user$_1$ is explicitly asking for hotel details) from the ancillary utterance by user$_2$. Even though the instructions clearly state that user$_1$’s state should be tracked, the presence of another speaker’s query creates an implicit context shift. As a result, the model may conflate or ignore the critical slot values it had previously captured.

In addition to missing values, multi-user input can also cause the model to produce erroneous slot types. Table~\ref{tab:error-analysis-2} demonstrates an example where the user requests a Mediterranean restaurant. In the single-user context, the LLM assigns the \textit{Restaurant-Food} slot as "mediterranean," which is correct. However, once user$_2$ joins the conversation expressing enthusiasm for a "Mediterranean experience" in the center of town, the model additionally (and incorrectly) infers a \textit{Restaurant-Area} slot value of "dontcare." This misprediction underscores how user$_2$’s mention of a location can trigger confusion about the user’s overall constraints or preferences. The second user’s statements appear to "override" or complicate the original request, leading the model to infer a new slot type (\textit{Restaurant-Area}) even though user$_1$ never expressed indifference about the area. Such discrepancies highlight a lack of robust multi-user grounding in the model: it fails to consistently distinguish separate user$_1$ requirements from user$_2$ remarks.

These errors collectively illustrate the challenges in scaling dialogue state tracking to multi-user scenarios. Both examples suggest that LLMs struggle to track the request of primary user and disentangle the utterance of other user. In multi-user dialogue applications (e.g., group travel planning or collaborative booking platforms), these behaviors could significantly degrade user experience and system reliability. Future efforts must address how best to isolate or merge multiple users’ goals to avoid contaminating one user’s state with another’s. Potential mitigation strategies include more explicit prompts that separate user$_1$’s goals from user$_2$’s utterance, or neural architectures that model each speaker’s state in parallel and then reconcile them at the system level. Overall, our results confirm that incorporating additional speakers in dialogue amplifies existing challenges in DST, highlighting the need for further research into robust, speaker-aware state tracking mechanisms.

\begin{table}[t!]
\centering
\small

\begin{tabularx}{\columnwidth}{|X|}
\hline
\textbf{Original MultiWOZ 2.1 w/o User$_2$ Utterance}\\
\hline
\textbf{User}: Do you have information about the Warkworth House?\\
\textit{[Hotel-Name: Warkworth House]}\\

\hline
\textbf{Extended MultiWOZ 2.1 w/ User$_2$ Utterance}\\
\hline
\textbf{User1}: Do you have information about the Warkworth House?\\
\textcolor{blue}{\textbf{User2}: I'm curious about the history of Warkworth House too!}\\
\textit{[Hotel-Name: \textcolor{red}{?}]}\\

\hline

\end{tabularx}
\caption{The error case where the LLM failed to extract correct slot value when user$_2$ utterance is introduced.}
\label{tab:error-analysis-1}
\end{table}

\begin{table}[t!]
\centering
\small

\begin{tabularx}{\columnwidth}{|X|}
\hline
\textbf{Original MultiWOZ 2.1 w/o User$_2$ Utterance}\\
\hline
\textbf{Agent}: I have curry garden for Indian in the centre of town, but no south indian.\\
\textbf{User}: What about one that serves mediterranean?\\
\textit{[Restaurant-Food: mediterranean]}\\

\hline
\textbf{Extended MultiWOZ 2.1 w/ User$_2$ Utterance}\\
\hline
\textbf{Agent}: I have curry garden for Indian in the centre of town, but no south indian. \\
\textbf{User1}: What about one that serves mediterranean?\\
\textcolor{blue}{\textbf{User2}: I'm in for a Mediterranean experience. Let's explore some top-notch options in the center together!}\\
\textit{[Restaurant-Food: mediterranean, \textcolor{red}{Restaurant-Area: dontcare}]}\\

\hline

\end{tabularx}
\caption{The error case where the LLM failed to predict correct slot type when user$_2$ utterance is introduced.}
\label{tab:error-analysis-2}
\end{table}

\newpage
\onecolumn
\section{Prompt Design}
\label{app:prompt}

\subsection{Prompt for Identification of Speech Act Type}
\label{app:prompt-identification}

Note that we shuffle the order of four options for each turn, to avoid biased identification \cite{lu2022fantastically}.

\begin{figure*}[h]
    \centering
    \lstset{basicstyle=\ttfamily\tiny, breaklines=true, postbreak=\mbox{\textcolor{red}{$\hookrightarrow$}\space}}
\begin{lstlisting}
Examine the previous dialogue history and choose the appropriate strategy for the utterance of User2 among the four options below. Considering the previous conversation, choose appropriate, unbiased representation for diversity.

Dialogue history so far:
{DIALOGUE HISTORY}

Current turn:
Agent: {AGENT UTTERANCE}
User1: {USER1 UTTERANCE}
User2: <New utterance to be generated>

Your task and requirements:
- Select one option among the four options mentioned below: 
    - Constatives : committing the speaker to something's being the case (answering, claiming, confirming, denying, disagreeing, stating)
    - Directives : attempts by the speaker to get the addressee to do something (apologizing, greeting, thanking, accepting an acknowledgment)
    - Commissives : committing the speaker to some future course of action (advising, asking, forbidding, inviting, ordering, requesting)
    - Acknowledgments : express the speaker's attitude regarding the hearer with respect to social action (promising, planning, vowing, betting, opposing)
- Output must be same words in prompt type option "Constatives", "Directives", "Commissives", "Acknowledgments".

\end{lstlisting}
\end{figure*}

\subsection{Prompt for Utterance Generation}
\label{app:prompt-generation}

\begin{figure*}[h]
    \centering
    \lstset{basicstyle=\ttfamily\tiny, breaklines=true, postbreak=\mbox{\textcolor{red}{$\hookrightarrow$}\space}}
\begin{lstlisting}
You are User2, a friend of User1. Your role is to casually communicate with User1 between Agent in less than 20 words following generation guide. User1 and User2 are talking to Agent.

Dialogue history so far:
{DIALOGUE HISTORY}

Current turn:
User1: {USER1 UTTERANCE}
User2: <New utterance to be generated>
Fixed slots: {SLOT VALUE OF CURRENT TURN}

Your task and requirements:

- Generate an appropriate utterance for User2 to respond to User1.
- Do not modify existing Current fixed slot and values or introduce new slot and values.
- Do not generate repeated utterances. (Ex. starting with 'Let's'or 'That sounds')
- Do not ask questions to User1. Only ask to Agent. Because the Agent will respond in the next turn.\n"
- Do not mention explitcitly word 'Agent'.\n"
- Note that you are in same circumstances with User1.
- Types of generated dialogues of User2: {UTTERANCE TYPE}

\end{lstlisting}
\end{figure*}

\subsection{Prompt for Validation of Generated Utterance}
\label{app:prompt-validation}

\begin{figure*}[h]
    \centering
    \lstset{basicstyle=\ttfamily\tiny, breaklines=true, postbreak=\mbox{\textcolor{red}{$\hookrightarrow$}\space}}
\begin{lstlisting}

You will be given a conversation sequence consisting of three parts:
1. Agent's response to previous Generated_User1 (`A`): {AGENT UTTERANCE}
2. User1's original utterance (`U1`): {USER1 UTTERANCE}
3. Generated_User2's transformed utterance (`U2`): {USER2 UTTERANCE}

Your task is to check the consistency of the sequence:
1. Check if User2's utterance (`U2`) is logically consistent with User1's original utterance (`U1`).
2. Verify that Generated_User2's (`U2`) slots match the current slot values from User1 (`U1`) slot: {SLOT VALUE OF CURRENT TURN}.
Output 'True' if both criteria 1 and 2 are met. Otherwise, output 'False'. Output: [True / False]

\end{lstlisting}
\end{figure*}

\newpage
\subsection{Prompt for Experiment}
\label{app:prompt-experiment}

Note that this is a slightly updated version of the prompt design suggested by previous study, considering the existence of user$_2$ \cite{heck2023chatgpt}.

\begin{figure*}[h]
    \centering
    \lstset{basicstyle=\ttfamily\tiny, breaklines=true, postbreak=\mbox{\textcolor{red}{$\hookrightarrow$}\space}}
\begin{lstlisting}
Consider the following list of concepts, called "slots" provided to you as a json list.

"slots": {   
    "taxi-leaveAt": "the departure time of the taxi",
    "taxi-destination": "the destination of the taxi",
    "taxi-departure": "the departure of the taxi",
    "taxi-arriveBy": "the arrival time of the taxi",
    "restaurant-book_people": "the amount of people to book the restaurant for",
    "restaurant-book_day": "the day for which to book the restaurant",
    "restaurant-book_time": "the time for which to book the restaurant",
    "restaurant-food": "the food type of the restaurant",
    "restaurant-pricerange": "the price range of the restaurant",
    "restaurant-name": "the name of the restaurant",
    "restaurant-area": "the location of the restaurant",
    "hotel-book_people": "the amount of people to book the hotel for",
    "hotel-book_day": "the day for which to book the hotel",
    "hotel-book_stay": "the amount of nights to book the hotel for",
    "hotel-name": "the name of the hotel",
    "hotel-area": "the location of the hotel",
    "hotel-parking": "does the hotel have parking",
    "hotel-pricerange": "the price range of the hotel",
    "hotel-stars": "the star rating of the hotel",
    "hotel-internet": "does the hotel have internet",
    "hotel-type": "the type of the hotel",
    "attraction-type": "the type of the attraction",
    "attraction-name": "the name of the attraction",
    "attraction-area": "the area of the attraction",
    "train-book_people": "the amount of people to book the train for",
    "train-leaveAt": "the departure time of the train",
    "train-destination": "the destination of the train",
    "train-day": "the day for which to book the train",
    "train-arriveBy": "the arrival time of the train",
    "train-departure": "the departure of the train"
}

Some "slots" can only take a value from predefined list:

"categorical": {
    "hotel-pricerange": ["cheap", "moderate", "expensive"],
    "hotel-area": ["north", "south", "east", "west", "centre"],
    "hotel-parking": ["yes", "no"],
    "hotel-internet": ["yes", "no"],
    "hotel-type": ["hotel", "guest house"],
    "restaurant-pricerange": ["cheap", "moderate", "expensive"],
    "restaurant-area": ["north", "south", "east", "west", "centre"],
    "attraction-area": ["north", "south", "east", "west", "centre"]
}

Now consider the following dialogue between more than two parties called the "system", "user1", and "user2".

Can you make answer as JSON ouput format which of the "slots" were updated by the "user1" in its latest response to the "system" refer to "slots" JSON list?

Present the updates in JSON format for each "slot" that was updated. There has no more than one update per "slot".
If no "slots" were updated, return an empty JSON list. If you encounter "slots" that were requested by the "user1" then fill them with "?". If a user does not seem to care about a discussed "slot" fill it with "dontcare".

\end{lstlisting}
\end{figure*}




\end{document}